\documentclass[10pt,twocolumn,letterpaper]{article}

\usepackage{iccv}
\usepackage{times}
\usepackage{epsfig}
\usepackage{graphicx}
\usepackage{amsmath}
\usepackage{amssymb}
\usepackage{subcaption}
\usepackage{enumitem}
\usepackage{paralist}
\setitemize{noitemsep,topsep=0pt,parsep=0pt,partopsep=0pt}
\usepackage[pagebackref=true,breaklinks=true,letterpaper=true,colorlinks,bookmarks=false]{hyperref}

\iccvfinalcopy 

\def\iccvPaperID{1601} 

\ificcvfinal\fi
\begin{document}

\title{Recognizing Semantic Features in Faces using Deep Learning}

\author{Amogh Gudi\\
VicarVision\\
Amsterdam\\
{\tt\small amogh@vicarvision.nl}
}

\maketitle

\begin{abstract}
   Human face constantly conveys information, both consciously and subconsciously. However, as basic as it is for humans to visually interpret this information, it is quite a big challenge for machines. Conventional semantic facial feature recognition and analysis techniques mostly lack robustness and suffer from high computation time. This paper aims to explore ways for machines to learn to interpret semantic information available in faces in an automated manner without requiring manual design of feature detectors, using the approach of Deep Learning. In this study, the effects of various factors and hyper-parameters of deep neural networks are investigated for an optimal network configuration that can accurately recognize semantic facial features like emotions, age, gender, ethnicity etc. Furthermore, the relation between the effect of high-level concepts on low level features is explored through the analysis of the similarities in low-level descriptors of different semantic features. This paper also demonstrates a novel idea of using a deep network to generate 3-D Active Appearance Models of faces from real-world 2-D images.
   
   For a more detailed report on this work, please see \cite{gudi2015recognizing}.
\end{abstract}
\section{Introduction}
\label{Introduction}
A picture is worth a thousand words, but how many words is the picture of a face worth? As humans, we make a number of conscious and subconscious evaluations of a person just by looking at their face. Identifying a person can have a defining influence on our conversation with them based on past experiences; estimating a person's age,  and making a judgement on their ethnicity, gender, etc. makes us sensitive to their culture and habits. We also often form opinions about that person (that are often highly prejudiced and wrong); we analyse his or her facial expressions to gauge their emotional state (e.g. happy, sad), and try to identify non-verbal communication messages that they intent to convey (e.g. love, threat). We use all of this information when interacting with each other. In fact, it has been argued that neonates, only 36 hours old, are able to interpret some very basic emotions from faces and form preferences \cite{farroni2007perception}. In older humans, this ability is highly developed and forms one of the most important skills for social and professional interactions. Indeed, it is hard to imagine expression of humour, love, appreciation, grief, enjoyment or regret without facial expressions.

\subsection{Semantic Features in Faces}
\label{semanticfeatures}

Predominantly, the following semantic features from the human face form the primary set of information that can be directly inferred (or roughly estimated) from faces (without contextual knowledge) by humans (apart from identity): expressed emotion, age, gender and ethnicity. In addition to these, certain other `add-on' features like presence of glasses and facial hair (beard, moustache), that are inherent properties of the face are also considered in this study.
\subsection{Objectives and Research Questions}
\label{thesisobjectives}
The main objective of this paper is to build and study a Deep Learning based solution to extract semantic features from images of faces. This sub-section describes the main questions that will be researched in the course of achieving this objective.

The design of a deep neural network for any particular task involves determining multiple configurations and parameters that can ensure that the network is well suited for the task at hand. Every such combination of hyper-parameters affects the output of the system differently. Therefore, one of the questions that will be researched as part of this paper is:
\textit{How could a deep learning technique adapt to the task of semantic facial feature recognition?} This question is closely followed by determining how different configurations, hyper-parameters (of the network), scale of the input, and addition of pre-processing steps affect the performance and accuracy of the system.


It is known that during the training of a multi-layered deep neural network, lower layers of the network learn to recognize low-level patterns (like edges), while the higher layers combine these low-level information to determine higher-level concepts. With respect to a deep network trained to recognize different semantic features in faces, the question that can be asked is: 
\textit{How are the high-level semantic descriptions related to their low-level feature descriptors? How are the low-level descriptors of deep networks, trained to classify different semantic features in faces, related to each other?}

Finally, there are several attributes in human faces whose semantics are not easily defined. For example, the contraction of specific facial muscles, or the locations of certain landmarks on the face may not lead to easily interpretable semantic information. However, such information can be useful for certain in-depth analysis (e.g., psychological research on human subconsciousness, lie-detection, etc.). A good representation of this information can be achieved through a 3-D Active Appearance Model \cite{van2005model} of the face. This leads to the following research question: 
\textit{Is a deep learning based method capable of generating 3-D Active Appearance Models of faces from a 2-D images?} 


\section{Related Work}
\label{RelatedWork}
\begin{figure}
	\centering
	\includegraphics[width=\linewidth]{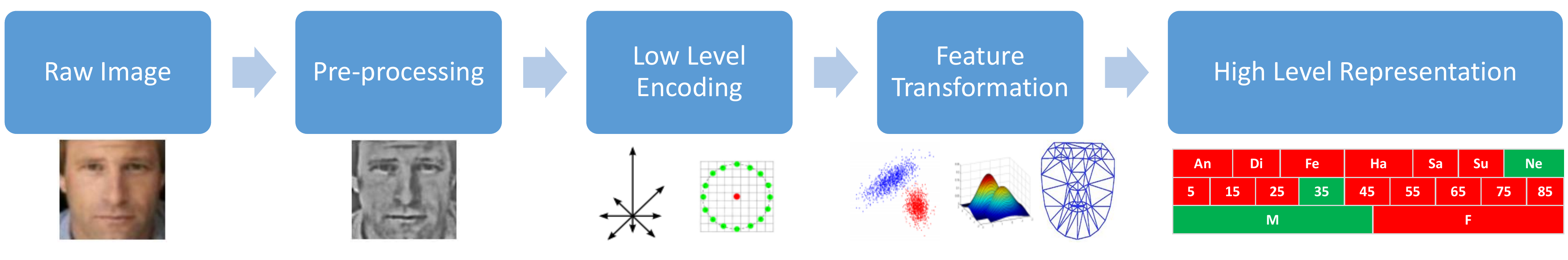}
	\caption[Typical conventional facial feature extraction pipeline]{Conventional facial feature extraction pipeline \cite{fan2014learning}.} 
    \label{fig:conventionalPipeline}
\end{figure}

The typical conventional approach for the task of facial feature recognition essentially follows the pipeline shown in Figure \ref{fig:conventionalPipeline}. Majority of the conventional and commercial facial analysis methods rely on the Facial Action Coding System (FACS) \cite{ekman1997face}, which involves identifying various facial muscles that can cause changes in physical facial appearance. \cite{van2005model} uses a model based approach called the Active Appearance Model \cite{cootes2004statistical} to classify emotion while building a 3-D model of the face that encodes over 500 facial landmarks from which facial muscular movements (Action Units, defined by the FACS) can be derived. The Active Appearance Model is generated using PCA \cite{sirovich1987low} directly on the pre-processed pixels, and is encoded as the deviation of a face from the average face. This model is then used to classify the emotions expressed by the face using a single layered neural network.

Some of the primary tasks within the field of computer vision are detection, tracking and classification. With the advent of deep learning, the state-of-the-art in all three of these tasks has considerably improved.
A successful demonstration of the capability of Deep Learning for the task of image classification/detection was done by Le et al. in \cite{le2011building}. 
Their results also showed that the detector can be sensitive to other non-target high-level categories which it encounters in the dataset (i.e. the unsupervised face detector also shows sensitivity to images of human bodies, cats and other high-level concepts). 
The study in  \cite{lecun2010convolutional} presents how important the pooling, rectification and contrast normalization steps can be in Deep Convolutional Neural Networks. Hinton and Srivastava successfully demonstrate further improvements in training by the use of dropouts in \cite{hinton2012improving, srivastava2013improving}. 
One of the most successful papers from 2012 showing the application of Deep Learning methods, specifically Deep Convolutional Networks, in image classification is \cite{krizhevsky2012imagenet} by Krizhevsky et al. Their work focuses on image recognition on ImageNet \cite{deng2009imagenet}. 
On the same dataset in 2013, work by Sermanet et al. \cite{sermanet2013overfeat} demonstrated an integrated solution by the use of Deep Convolutional Neural Networks for all the three tasks of detection, localization and classification. This work attained the state-of-the-art in the Classification+Localization task. Baccouche et al. in \cite{baccouche2011sequential} demonstrated the use of 3-Dimensional Convolutional Neural Networks in combination with Recurrent Neural Networks for human-action classification in videos, thus making use of spatial as well as temporal information in the frame sequences to generate state-of-the-art results.

In the context of this paper, we focus on the task of facial feature recognition. There is a large body of research dedicated to this problem, and deep learning has emerged as a highly promising approach in solving such tasks. A recent study  \cite{taigman2013deepface} has shown near-human performance using deep networks in the task of recognizing the identity of a person from faces. With the use of preprocessing steps like face alignment and frontalization, and the use of a very large dataset, a robust and invariant classifier is produced that sets the state-of-the-art in the Labelled Faces in the Wild dataset \cite{huanglabeled}. This work utilises a modified version of deep convolutional networks, with certain convolutional layers using unshared weights (while regular convolutional layers share weights).

In the task of emotion recognition from faces, Tang's \cite{tang2013deep} sets the state-of-the-art on the Facial Expression Recognition Challenge (FERC) dataset. This is achieved by implementing a two stage network: a convolutional network trained in a supervised manner on the first stage, and a Support Vector Machine as the second stage trained on the output of the first stage. Recent work by Kahou et al. in \cite{kahou2015emonets} successfully demonstrates a multi-modal deep learning based framework for emotion recognition in videos. 


\section{The Experimental Set-up}
\label{MyModel}

\subsection{The Task}
The task of recognizing semantic features in faces is essentially an umbrella term for deciphering information encoded in faces in general, both apparent and not-so-apparent. Thus, it can be viewed as a task of extracting information from images with the prior knowledge that the images represent human faces. A list of such information encoded in images of human faces is provided in sub-section \ref{semanticfeatures}.

\subsection{The Datasets}

\begin{figure}
	\centering            
	\includegraphics[width=\linewidth]{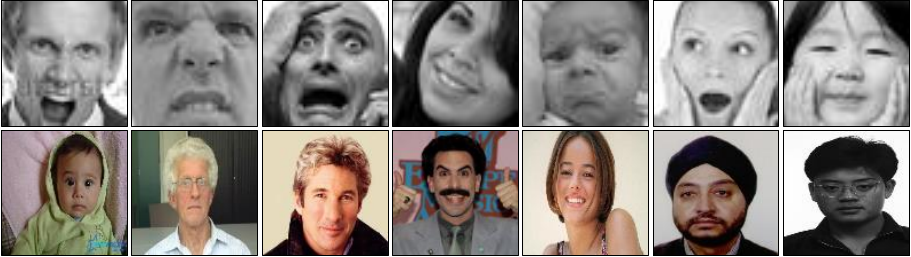}
	\caption[Dataset Examples]{Dataset examples (Top Row: FERC Dataset, Bottom Row: VV Dataset).}
	\label{fig:DatasetExamples}
\end{figure}

\begin{figure}
\begin{subfigure}{\linewidth}
	\centering
	\includegraphics[width=\linewidth]{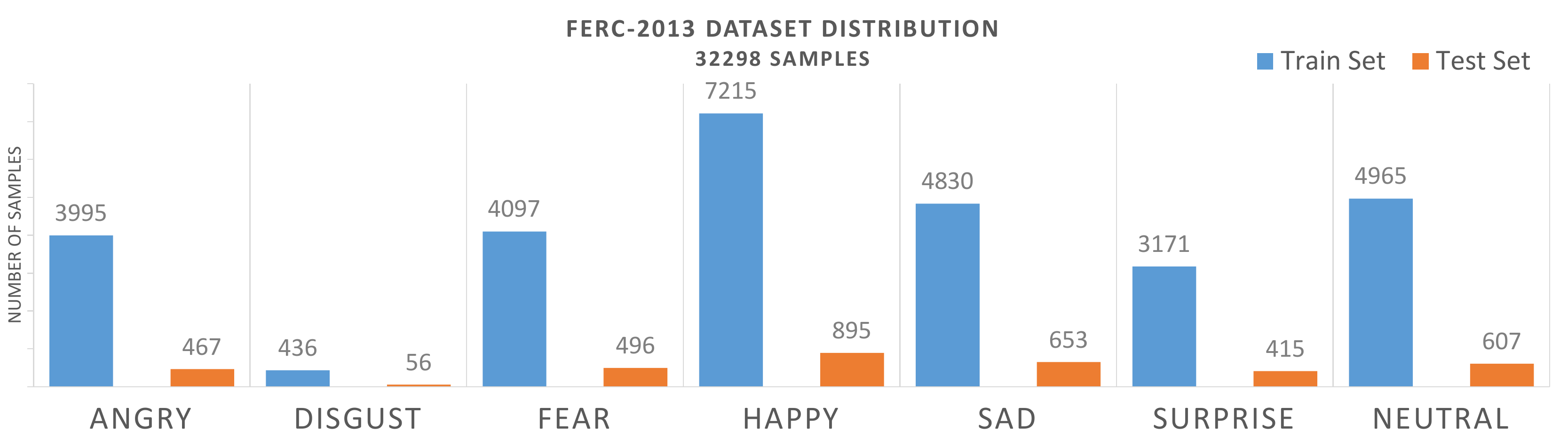}
	\caption{FERC Dataset.}
    \label{fig:FERCHistogram}
\end{subfigure}
\\
\begin{subfigure}{\linewidth}
	\centering
	\includegraphics[width=\linewidth]{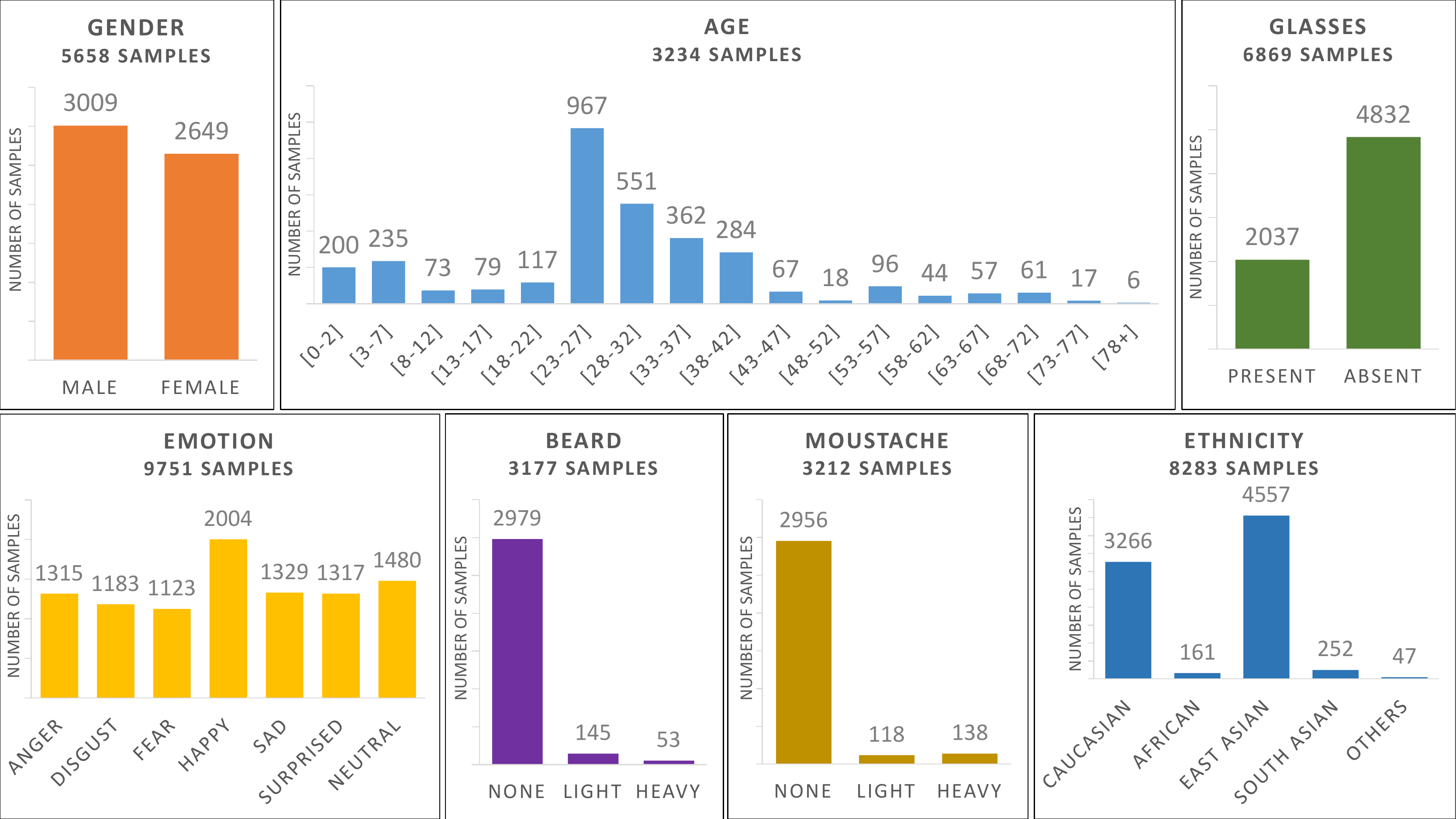}
	\caption{VV Dataset.}
    \label{fig:VVDatasetHistogram}
\end{subfigure}
    \caption[Dataset Histograms]{Distribution of classes in the datasets.}
    \label{fig:DatasetHistograms}
\end{figure}

Two datasets were used for the training and testing of the network in this paper: The emotion-annotated dataset from the Facial Expression Recognition Challenge \footnote{http://www.kaggle.com/c/challenges-in-representation-learning-facial-expression-recognition-challenge/data}, and the multi-annotated private dataset from VicarVision\footnote{http://www.vicarvision.nl/} (hereby referred to as the VV dataset). Examples and the statistics of the datasets are shown in figure \ref{fig:DatasetExamples} and \ref{fig:DatasetHistograms}.

\subsection{Pre-processing Steps}
\label{preprocessing}
It can be difficult for the deep network to be able to handle high variations in the pose of faces, and in lighting conditions of the image. Thus, it becomes necessary to pre-process the input so as to make the faces more uniform.

\begin{figure}
	\centering            
	\includegraphics[width=\linewidth]{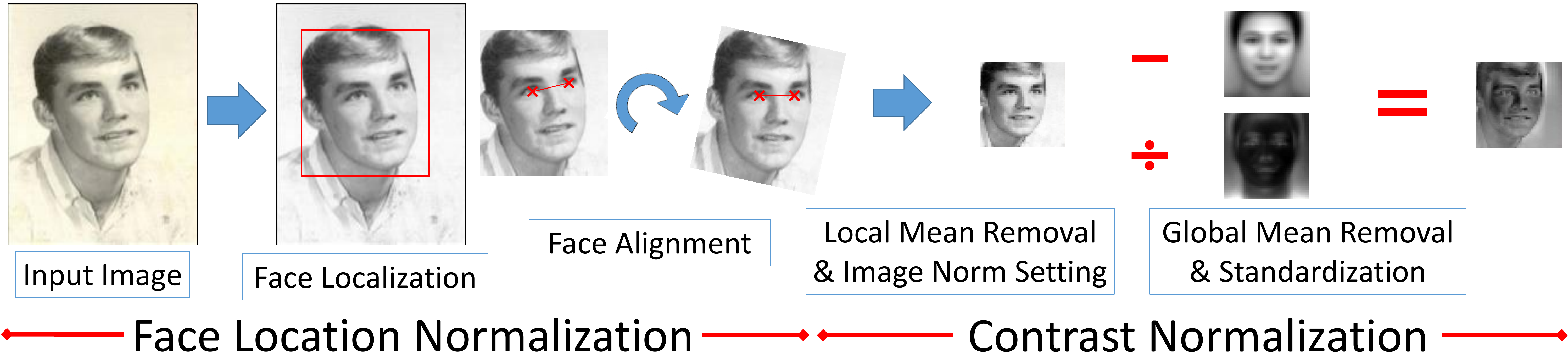}
	\caption[Pre-processing pipeline]{The pre-processing pipeline.}
    \label{fig:prepro}
\end{figure}

The pre-processing steps can be divided into two parts as they seek to minimize two distinct properties of the input image: the variation in location and pose of the face, and the variations in lighting conditions and contrast.
The basic pipeline of the pre-processing steps is illustrated in Figure \ref{fig:prepro}.

\noindent\textbf{Face Location Normalization}: 
	\begin{compactitem}
	\item Find faces in the image using the a face detection algorithm (specifically, the Viola/Jones face detection algorithm 
	\cite{viola2001rapid}) and extract the face-crop. 
	\item Perform in-plane rotate so as to remove tile of the faces in the X-Y plane. 
	\item Resize the image such that the approximate scale of the face is constant. This is done by ensuring that the distance between the two eyes in the faces is constant.
	\end{compactitem}

\noindent\textbf{Global Contrast Normalization}
	\begin{compactitem}
	\item For each image, subtract the local mean of all pixel values from the image. 
	\item Set the image norm to be equal to 100. 
	\item For each pixel in each image, subtract the global mean of pixels at that location throughout the dataset (the train set), and divide by the standard deviation.
	
	\end{compactitem}

\subsection{Training Method}
\label{trainingmethod}
Throughout the experiments mentioned in this paper, training of the network is done using stochastic gradient descent with momentum in mini-batch mode, with batches of 100 data samples. Negative log-likelihood is used as the objective function. Learning rate for the training is initialized to 0.0025, and is linearly decreased to 0.001 over 50 epochs of training. 

Training is evaluated using a validation set, which is roughly 10\% of the size of the total dataset (train set + valid set + test set). Stopping criteria of the network training is based on the misclassification rate/mean squared error on the validation set.
The network is tested on a test set which also contains about 10\% of the data samples in the dataset.

\section{Experiments and Results}
\label{ExperimentsAndResults}
In this section, a description of all experiments is given, and results of the performance of the network on various test sets are provided. All experiments have been performed on Nvdia GTX 760\footnote{http://www.nvidia.com/gtx-700-graphics-cards/gtx-760/}. The \textit{theano} framework \cite{theano} based \textit{pylearn2} library \cite{pylearn2} has been primarily used for these experiments.

\subsection{Experiments on the FERC Dataset}
This sub-section describes the experiments conducted for the emotion recognition task on the FERC Dataset under different network configurations as well as training parameters. As a baseline, it is useful to note that a random classifier produces an accuracy of 14.3\%, and a single-layer softmax regression model gives 28.16\% accuracy. 

\subsubsection{Best Performing Deep Network}
\label{bestnetwork}
\begin{figure}
	\centering
	\includegraphics[width=\linewidth]{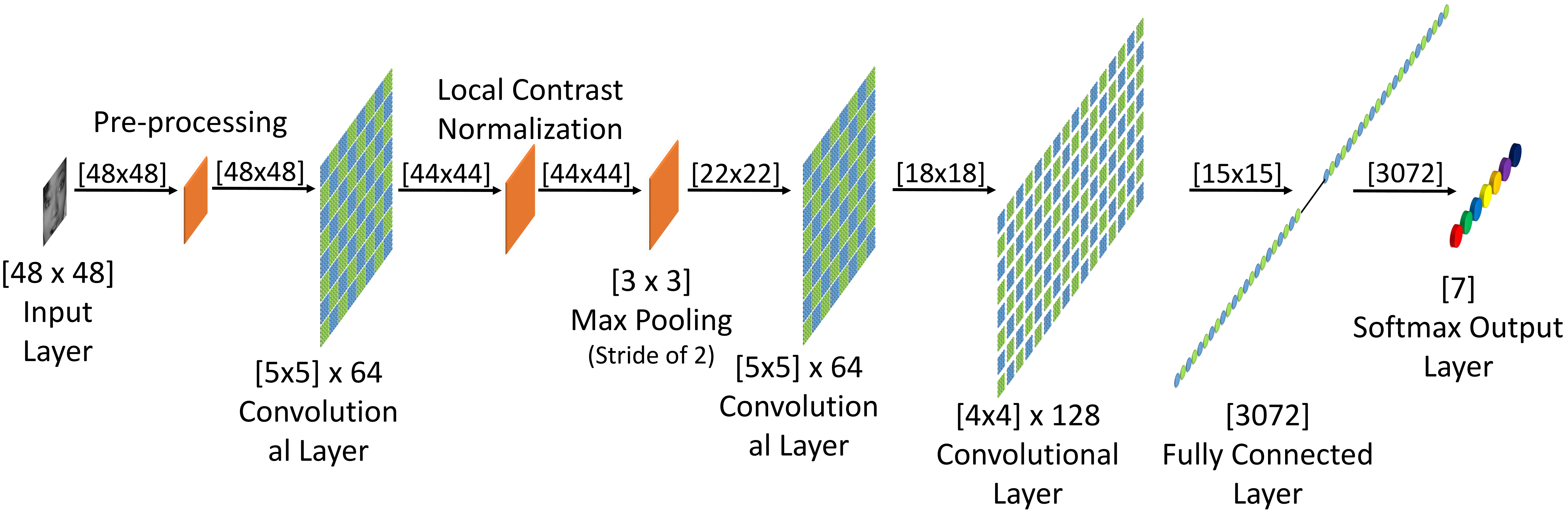}
	\caption[Deep network architecture]{Architecture of the deep network.}
    \label{fig:convnet}
\end{figure}

The architecture and hyper-parameters for this network are obtained on the basis of empirical results described later.
\\The input image in the form of $48 \times 48$ grayscale pixels arranged in a 2D matrix is fed to the first hidden layer of the network: A convolutional layer with a kernel size of $5 \times 5$ having a stride of 1 both dimensions. The number of parallel feature-maps in this layer is $64$. The $44 \times 44$ output image produced by this layer is passed to a local contrast normalization and a max-pooling layer \cite{lecun2010convolutional} of kernel size $3 \times 3$ with a stride of 2 in each dimension (selection based on previous work in \cite{krizhevsky2012imagenet, tang2013deep}). This results in a sub-sampling factor of $1/2$, and hence the resulting image is of size $22 \times 22$. The second hidden layer is also a 64 feature-map convolutional layer with a kernel size of $5 \times 5$ (and stride 1). The output of this layer is a $18 \times 18$ pixel image, and this feeds directly into the third hidden layer of the network, which is a convolutional layer of 128 feature maps with a kernel size of $4 \times 4$ (and stride 1). Finally, the output of this layer, which is of dimension $15 \times 15$, is fed into the last hidden layer of the network, which is a fully connected linear layer with 3072 neurons. Dropout is applied to this fully connected layer, with a dropout probability of 0.2. The output of this layer is connected to the output layer, which is composed of 7 neurons, each representing one class label. Because this dataset has mutually exclusive emotional expression labels, a softmax operation is performed on the output of these 7 neurons and the class with the highest activation is chosen. All layers in the network are made up of ReLu units/neurons \cite{nair2010rectified}. This architecture is illustrated in Figure \ref{fig:convnet}. The network is trained using stochastic gradient descent, as described in Section \ref{trainingmethod}. 

\begin{figure}
	\centering
     \begin{subfigure}{0.495\linewidth}
        \centering
    	\includegraphics[width=\linewidth]{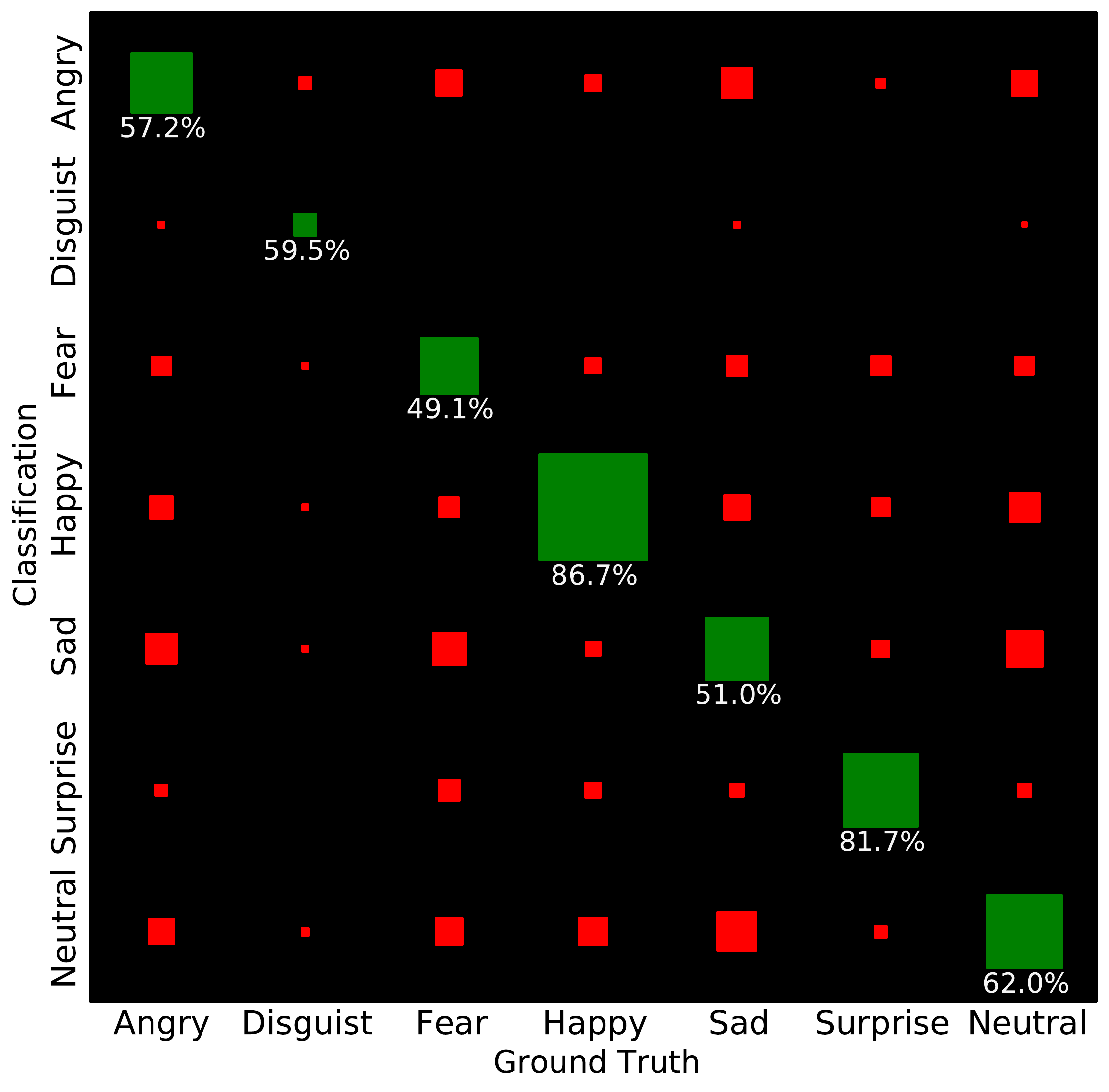}
    	\caption{FERC Test set classification confusion matrix.}
    	\label{fig:FERCFilAliClassification}
    	\end{subfigure}
        \begin{subfigure}{0.495\linewidth}
        \centering
    	\includegraphics[width=\linewidth]{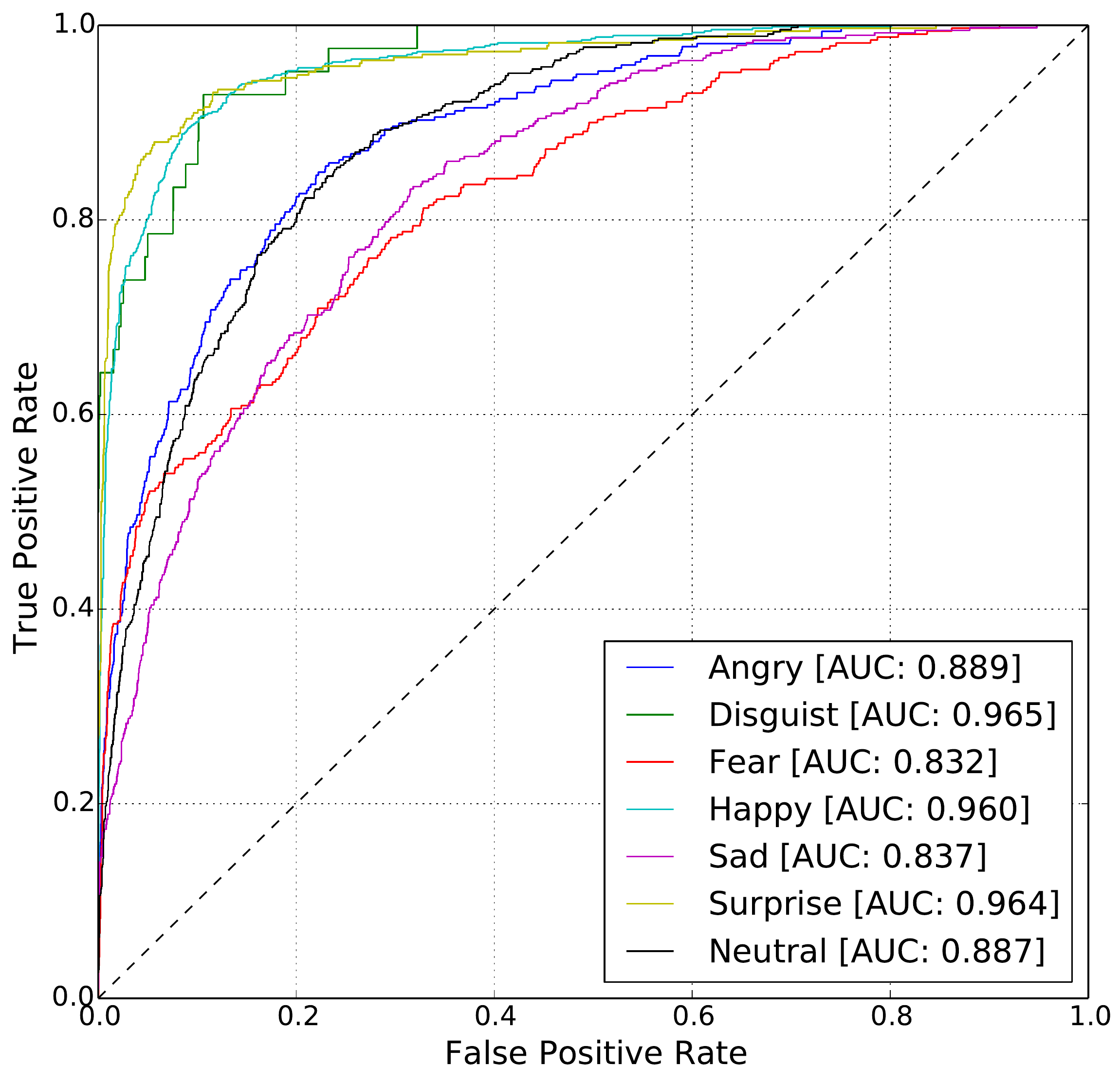}
    	\caption{FERC test set ROC curves (one-vs-all classes).}
    	\label{fig:FERCFilAliROC}
        \end{subfigure}%
        \caption[Best performance on the FERC test set]{Best performance on the FERC Test set: The total classification accuracy of the network was 67.12\%, with the average precision per class being 59.6\%.}
        \label{fig:FERCResults}
\end{figure}

The performance of this network on the test set can be viewed in Figure \ref{fig:FERCResults}. The network is able to correctly classify 67.12\% of the test samples, maintaining an average precision per class of 59.6\%. It can be seen that the network has over 50\% precision for all classes except fear (which is 49.1\%). This could be because the visual appearance of a face expressing fear varies considerably for different people, and is often confused with surprise and sadness. It can be noted from ROC plot that disgust, happy and surprised show very good discrimination qualities, despite having relatively very few samples in the training set \ref{fig:FERCHistogram}. This could be due to a relatively large difference in the visual appearance of disgusted faces as compared to other emotions. 

\vspace{-3ex}
\paragraph{Comparison with Sate-of-the-art}
The state-of-the-art results on the complete FERC test set 
is 69.4\% total classification accuracy (as reported by Charlie Tang in \cite{tang2013deep}). This is achieved by a network with similar architecture, but with the absence of the face location normalization pre-processing step and the sofmax layer, along with the addition of a 2nd stage SVM classifier. 

\subsubsection{Experiments with Network Size}
In this experiment, the width (average number of neurons per layer) of a convolutional network was altered by changing the number of feature maps in the network (while keeping the size of the convolutional kernel fixed). The depth (number of layers) of the network was altered by the addition or removal of a convolutional layer in the network, while keeping the fully connected layer always at the last position.

\begin{figure}

\centering
     \begin{subfigure}{0.6\linewidth}
        	\centering
        	\includegraphics[width=\linewidth]{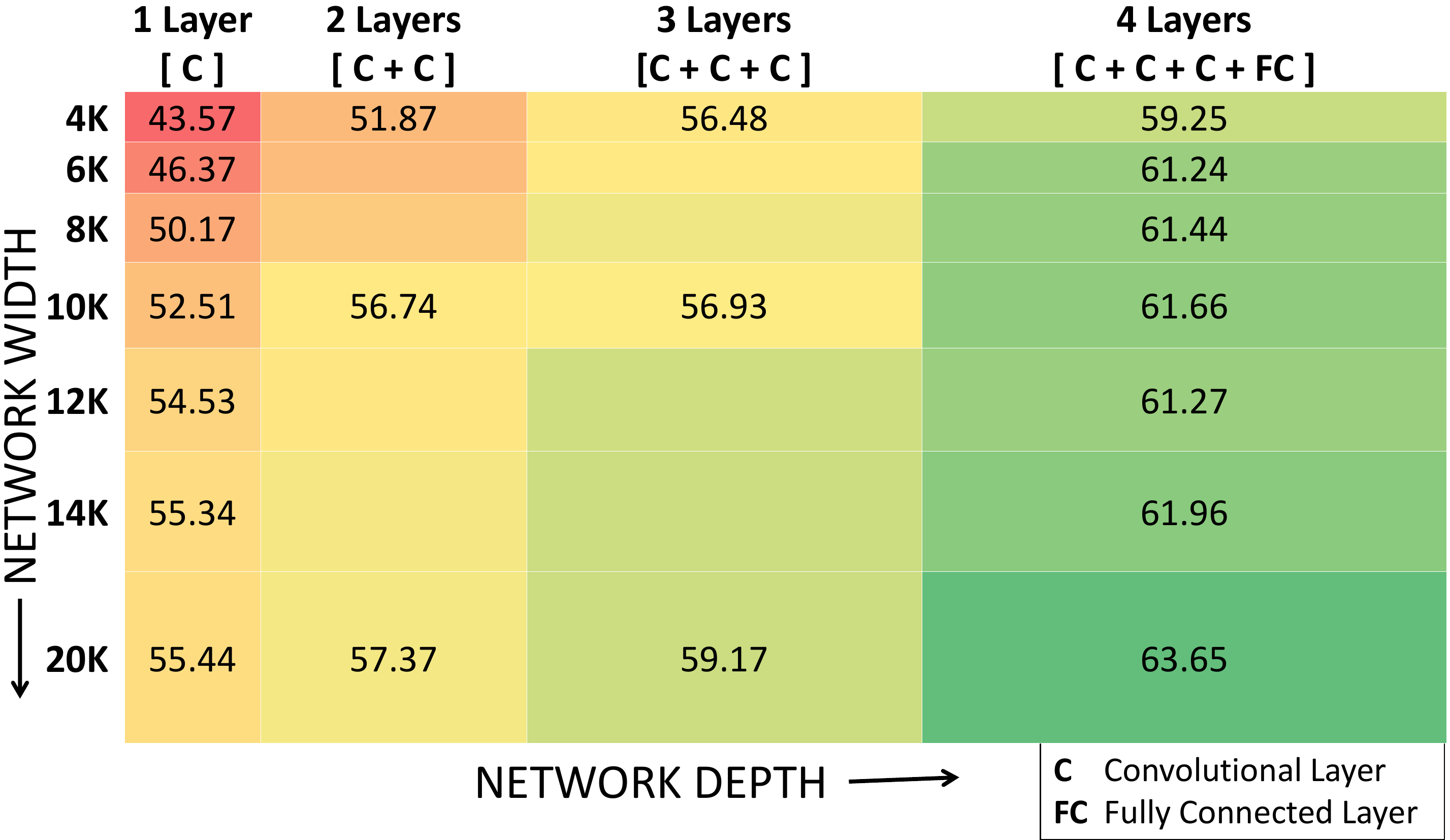}
        	\label{fig:depthvswidth_table}
        	\caption{Classification accuracies of networks with varying depth and width.}
        	
        \end{subfigure}
        \begin{subfigure}{0.39\linewidth}
        	\centering
        	\includegraphics[width = \linewidth]{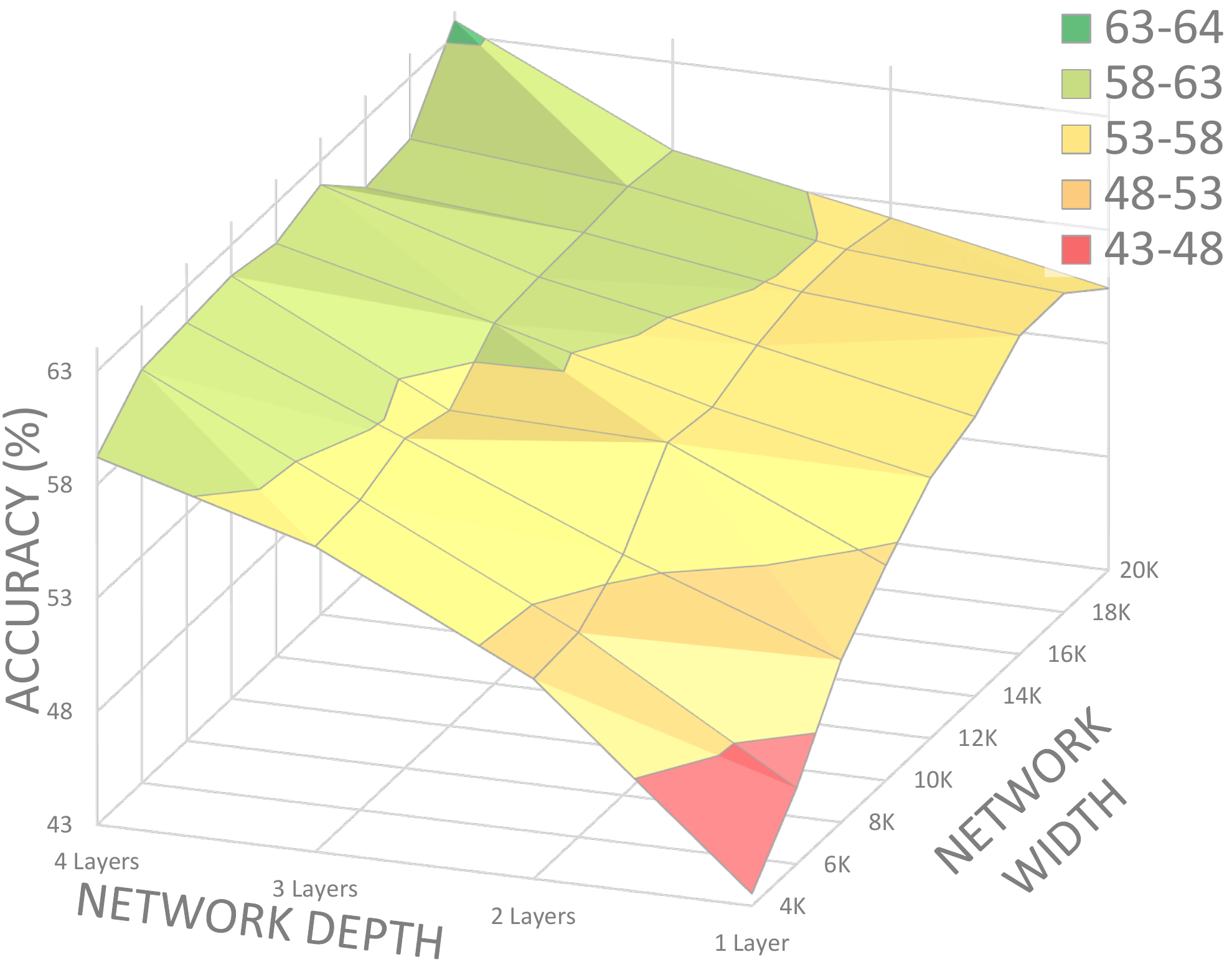}
    	      	\label{fig:depthvswidth_surf}
    	      	\caption{Surface plot of classification accuracy in the network depth vs width space.} 
        \end{subfigure}%
        \caption[Network performance in depth vs width space]{Network performance in terms of accuracy with varying depth vs witdh.}
        \label{fig:DepthvsWidth}
\end{figure}

 Figure \ref{fig:DepthvsWidth} shows a heat-map table with accuracy for different depths and widths of the network. As can be seen, lower depth and width gives the lowest accuracy, while higher depth and width provide the highest accuracy. This suggests the very intuitive fact that larger the network, the better the performance. Closer examination of these results and the surface plot in figure \ref{fig:DepthvsWidth} also show that the depth of the network has a higher impact when compared to the width of the network. However, after 3 layers, this impact seems to get smaller. Similar effects can be seen with the width of the network. 


\subsubsection{Experiments with LCN and Pooling}

\begin{figure}
\centering
     \begin{subfigure}{0.55\linewidth}
        	\centering
        	\includegraphics[width=\linewidth]{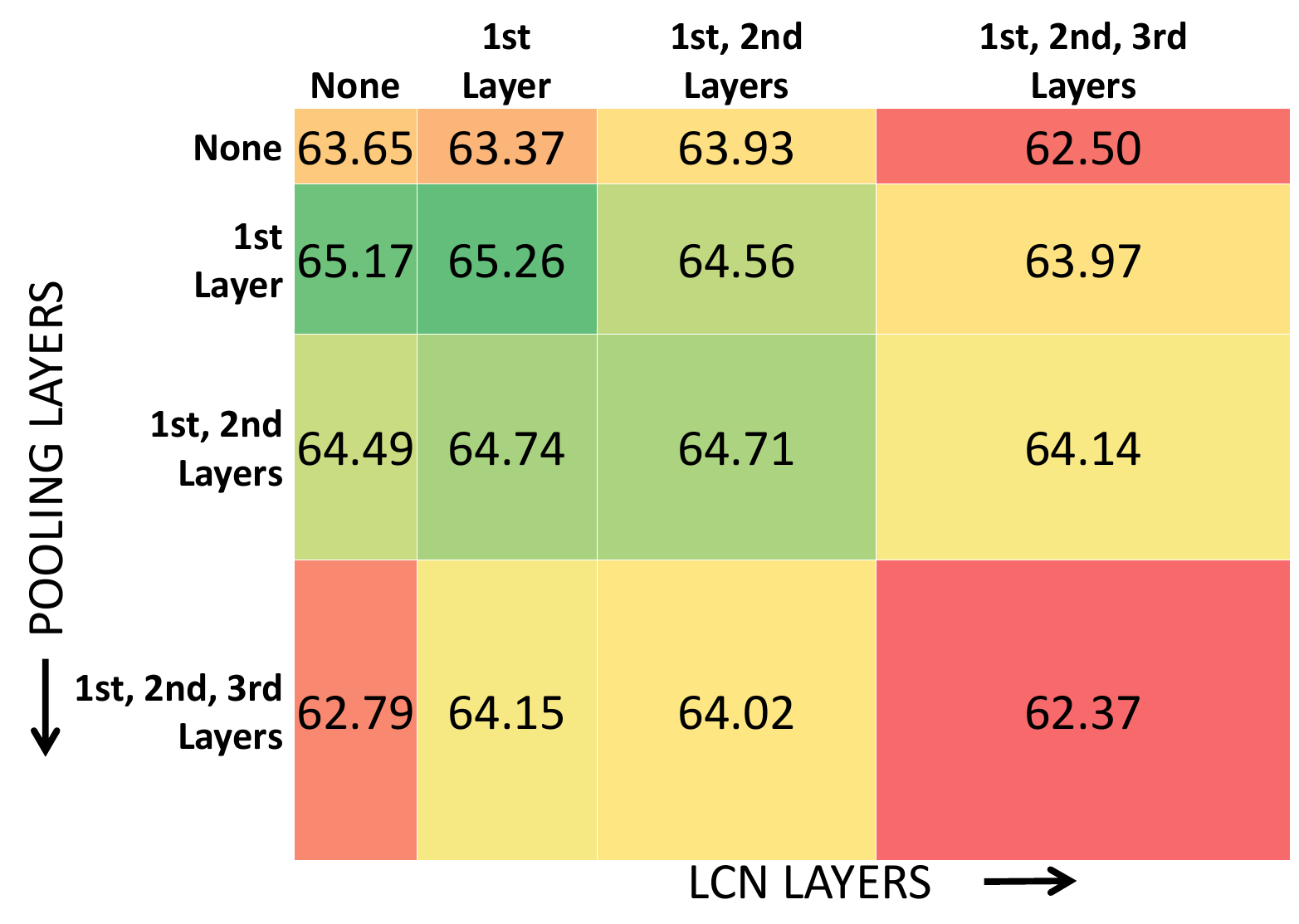}
        	\label{fig:lcnvspooling_table}
        	\caption{Classification accuracies of networks with varying applications of LCN and max-pooling.} 
        \end{subfigure}
        \begin{subfigure}{0.44\linewidth}
        	\centering
        	\includegraphics[width = \linewidth]{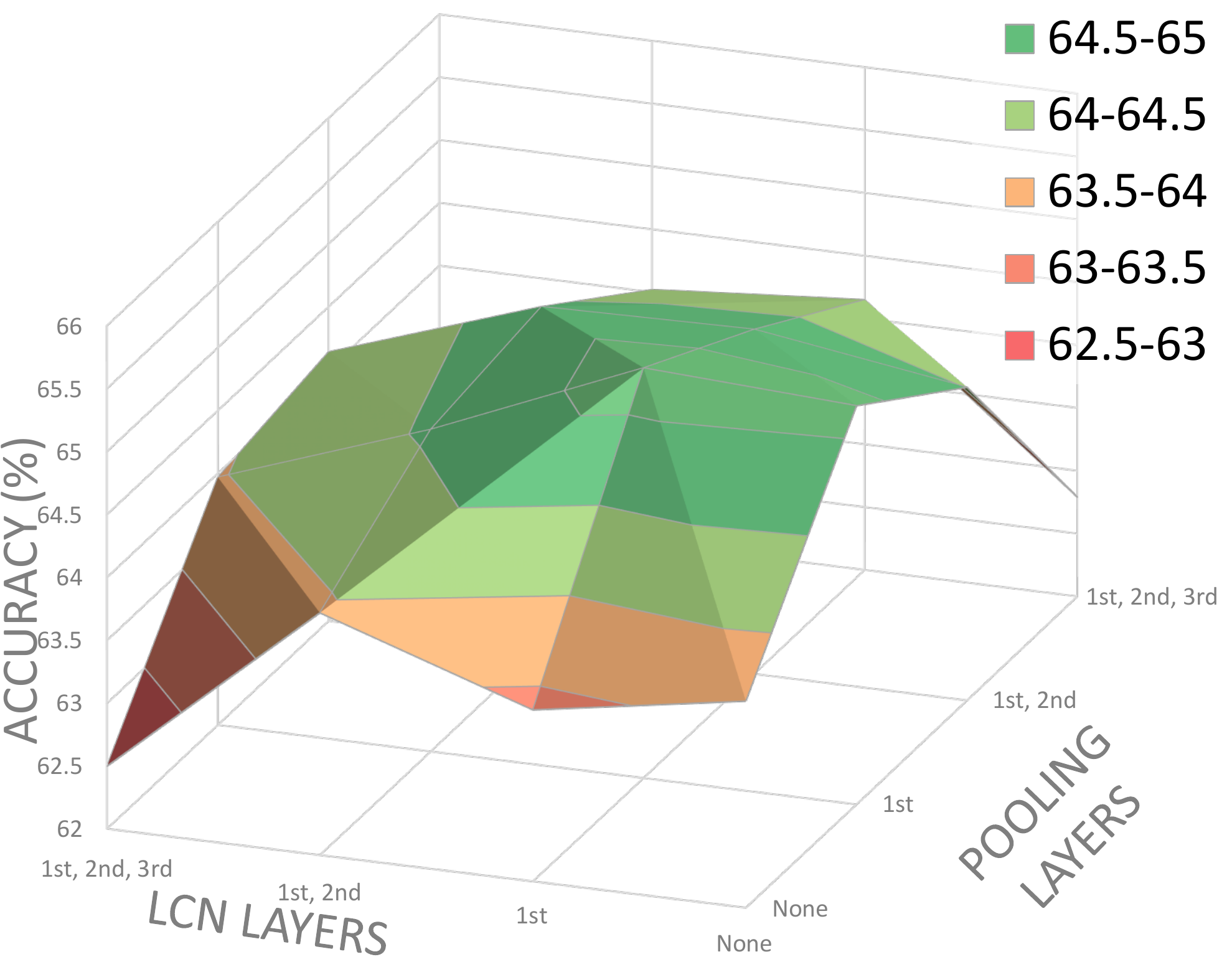}
    	      	\label{fig:lcnvspooling_surf}
    	      	\caption{Surface plot of classification accuracy in the application of pooling vs LCN space.}
        \end{subfigure}%
        \caption[Network performance in pooling vs LCN space]{Network performance in terms of accuracy with varying applications of pooling vs local contrast normalization.}
        \label{fig:LCNvsPooling}
\end{figure}

This experiment was conducted in order to determine the closest-to-optimal combination of Local Contrast Normalization (LCN) and max-pooling within the neural network layers. Max-pooling essentially results in a non-linear down-sampling step, introducing translation invariance and reducing computational complexity. 
Local contrast normalization is another well-used step in designing deep architectures, which ensures competition among the activations of nearby neurons by normalizing them locally with respect to each other. 

For the purpose of this experiment, a network similar to the one shown in figure \ref{fig:convnet}, without the pooling and LCN layers, is considered as the baseline. Max-pooling (with a kernel size of $3 \times 3$ in strides of $2$) and LCN are then applied at three locations within the network: at the outputs of the first, second and third convolutional layers. 

The results of this experiment can be seen in Figure \ref{fig:LCNvsPooling}. 
Max-pooling the outputs of the first convolution layer, after applying LCN, gives the best results. Simply applying LCN without pooling the outputs degrades the results, and this might be because in the absence of pooling, normalizing the outputs locally leads to extra emphasis on certain non-informative activations (which otherwise would not have propagated beyond the pooling stage).
Also, applying pooling to the network is relatively more advantageous in the starting layers of the network. 
This could be attributed to the fact that activations of deeper layers represent more information than the activations of starting layers, and hence down-sampling these outputs lead to loss of useful information. 

\subsubsection{Experiments with Dropout}

Dropout essentially means randomly omitting the neurons of a layer by a certain probability. Dropout is an important recent improvement for neural networks. It works equivalent to adding random noise to the representation (randomly setting outputs of neurons to zero), or performing model averaging, and this helps reduce overfitting \cite{hinton2012improving,srivastava2013improving}.

In this experiment, the neural network described in Figure \ref{fig:convnet} is used, and dropout is applied on its layers during the training phase with varying magnitudes. 

\begin{figure}
\centering
     \begin{subfigure}{0.55\linewidth}
        	\centering
        	\includegraphics[width=\linewidth]{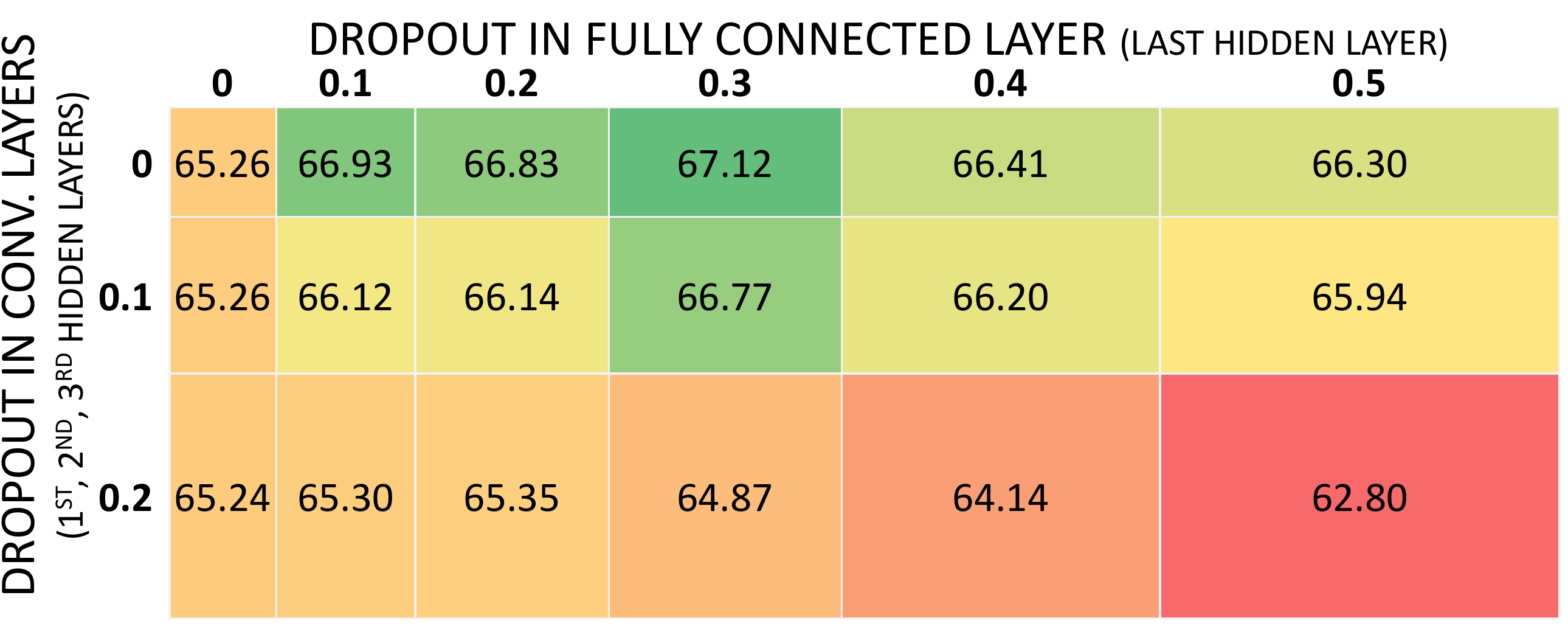}
        	\label{fig:dropout_table}
        	\caption{Classification Accuracies of Networks with varying dropouts in the convolutional and fully connected layers. The dimensions of each cell are proportional to the dropout probability.}
        \end{subfigure}
        \begin{subfigure}{0.44\linewidth}
        	\centering
        	\includegraphics[width = \linewidth]{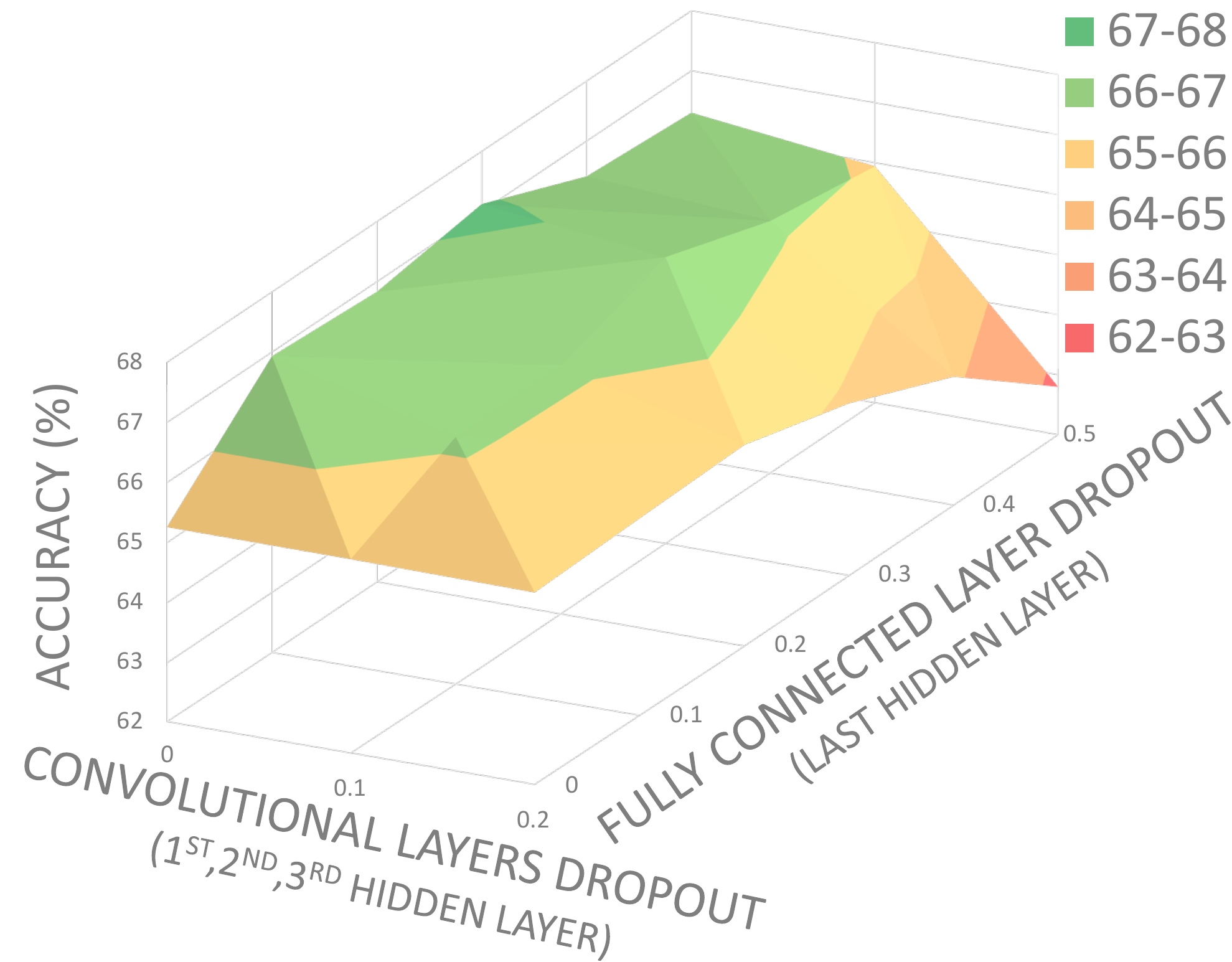}
    	      	\label{fig:dropout_surf}
    	      	\caption{Surface plot of classification accuracy in the network depth vs width space.}
        \end{subfigure}%
        \caption[Network performance in dropout space]{Network performance in terms of accuracy with varying magnitudes of the dropout probability in the final fully connected layer vs the convolutional layers.}
        \label{fig:dropoutvsdropout}
\end{figure}

The results of this experiment are shown in Figure \ref{fig:dropoutvsdropout}. 
It can be observed that a fully-connected layer with a dropout probability of $0.3$ gives the best performance, and applying any dropout to the convolutional layers only results in reduction in performance. These results support the optimzsed network architecture used in \cite{krizhevsky2012imagenet} where the best performance was obtained by using dropouts only on fully connected layers. This can be attributed to the fact that fully-connected layers are more prone to overfitting, while the additional noise caused by dropouts could be adversely affecting the convolutional feature detector.

\subsection{Experiments on the VV Dataset}
In this sub-section, the details and results of the experiments conducted on the VV Dataset are provided. 
This time the experiments do not focus on network optimisation. 
The network used for training and testing for recognition of various features in the dataset has the same architecture as defined in Section \ref{bestnetwork}, which is the most optimized network for the FERC dataset. In all the experiments that follow, the training of the network was done as described in Section \ref{trainingmethod}. 

\subsubsection{Emotion Classification}

The task of emotion classification on the VV Dataset is very similar to that on the FERC dataset. However, a key difference between the two datasets is that all the emotion classes are more uniformly distributed, as can be seen in Figure \ref{fig:VVDatasetHistogram}. 

The network produced a total classification accuracy of 66.56\%, while the average precision was 65.64\%. 
The performance of the network is quite similar to the one seen for the FERC dataset. The average precision score is closer to the total classification accuracy due to the uniform class distribution in the dataset. The ROC curves of Happy and Neutral show that they are the best-learned classification categories of the network, although all the other labels also have a decent amount of area under their curves ($>$0.9).

\vspace{-3ex}
\paragraph{Experiments with Input Image Resolution}

\begin{figure}
	\centering
	\includegraphics[width=\linewidth]{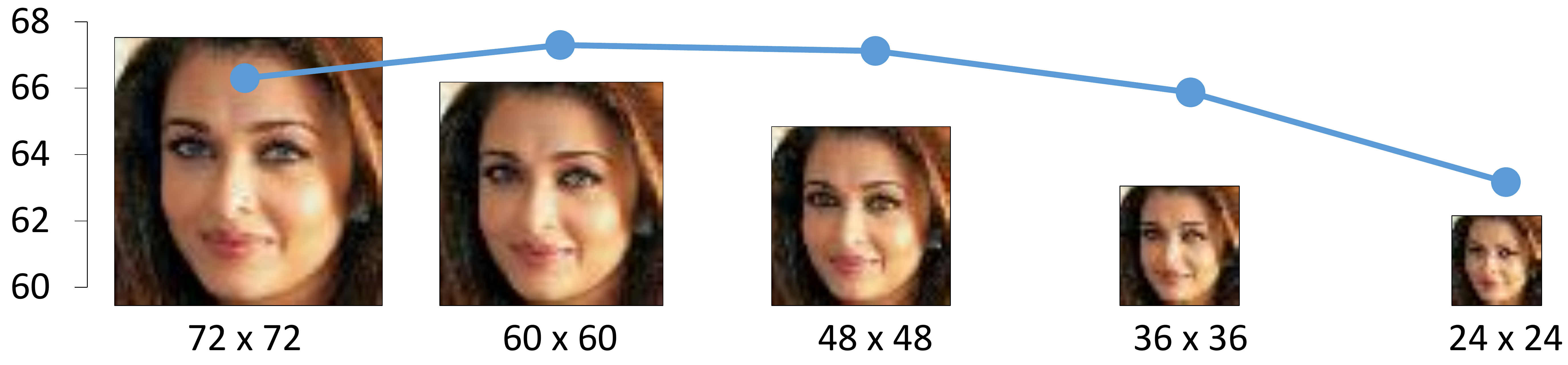}
	\caption[Experiments with Input image sizes]{Classification accuracies for various input image sizes provided to the network.} 
    \label{fig:inputsize}
\end{figure}

The results of ethe experiments using different image sizes can be observed in Figure \ref{fig:inputsize}.The performance of the network is about the same for image sizes of $60 \times 60$ and $48 \times 48$, and reduces smoothly for smaller image sizes. 
The performance of the network drops when image size in increased to $72 \times 72$ pixels, and this could be due to the fact that we are using a constant $5 \times 5$ sized convolution kernel, and not scaling it up with the input image size (due to limitations of computational resources). 

\vspace{-3ex}
\paragraph{Experiments with Pre-processing}
It is found that the global contrast normalization pre-processing step gives a performance boost of around 3\% to the classification accuracy, while the face alignment step improves the accuracy by roughly 5\%. 

\subsubsection{Age Classification}
In this experiment, the age annotations in the VV dataset are considered, which contain one of 17 exclusive age labels for each image. The age labels represent 5 year age intervals around ages that are multiples of 5 (except for the range [0-2]). To accommodate this, the final softmax layer of our network architecture is set to have 17 neurons, one for each age class. The network is trained as usual (as explained in sub-section \ref{trainingmethod}).

\begin{figure}
	\centering
    \begin{subfigure}{0.495\linewidth}
        \centering
    	\includegraphics[trim = 0mm 0mm 0mm 13.4mm, clip,width=\linewidth]{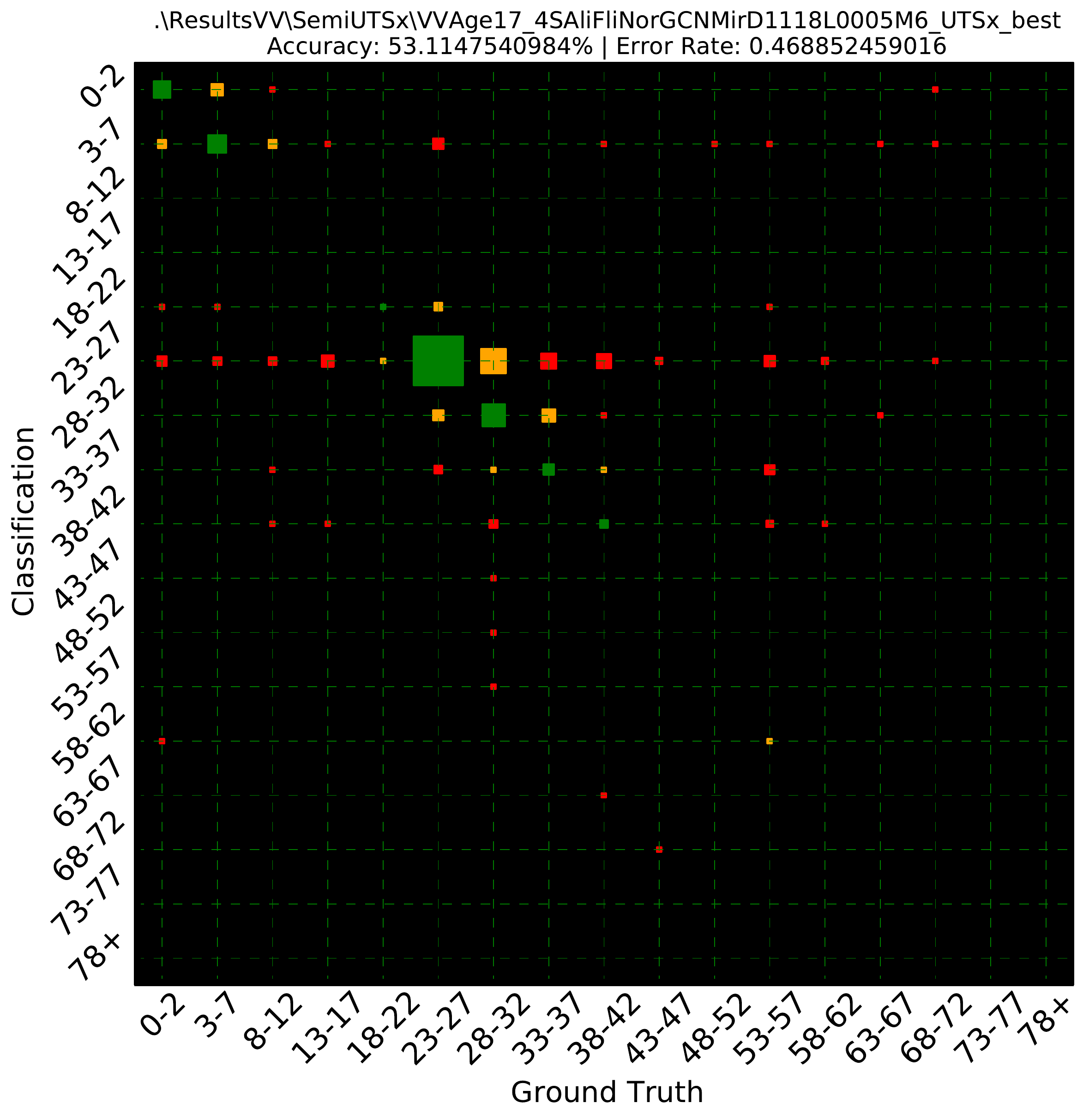}
    	\caption{VV dataset age classification confusion matrix.}
    	\label{fig:VVAgeClassification}
    \end{subfigure}
    \begin{subfigure}{0.495\linewidth}
	    \centering
   		\includegraphics[width=\linewidth]{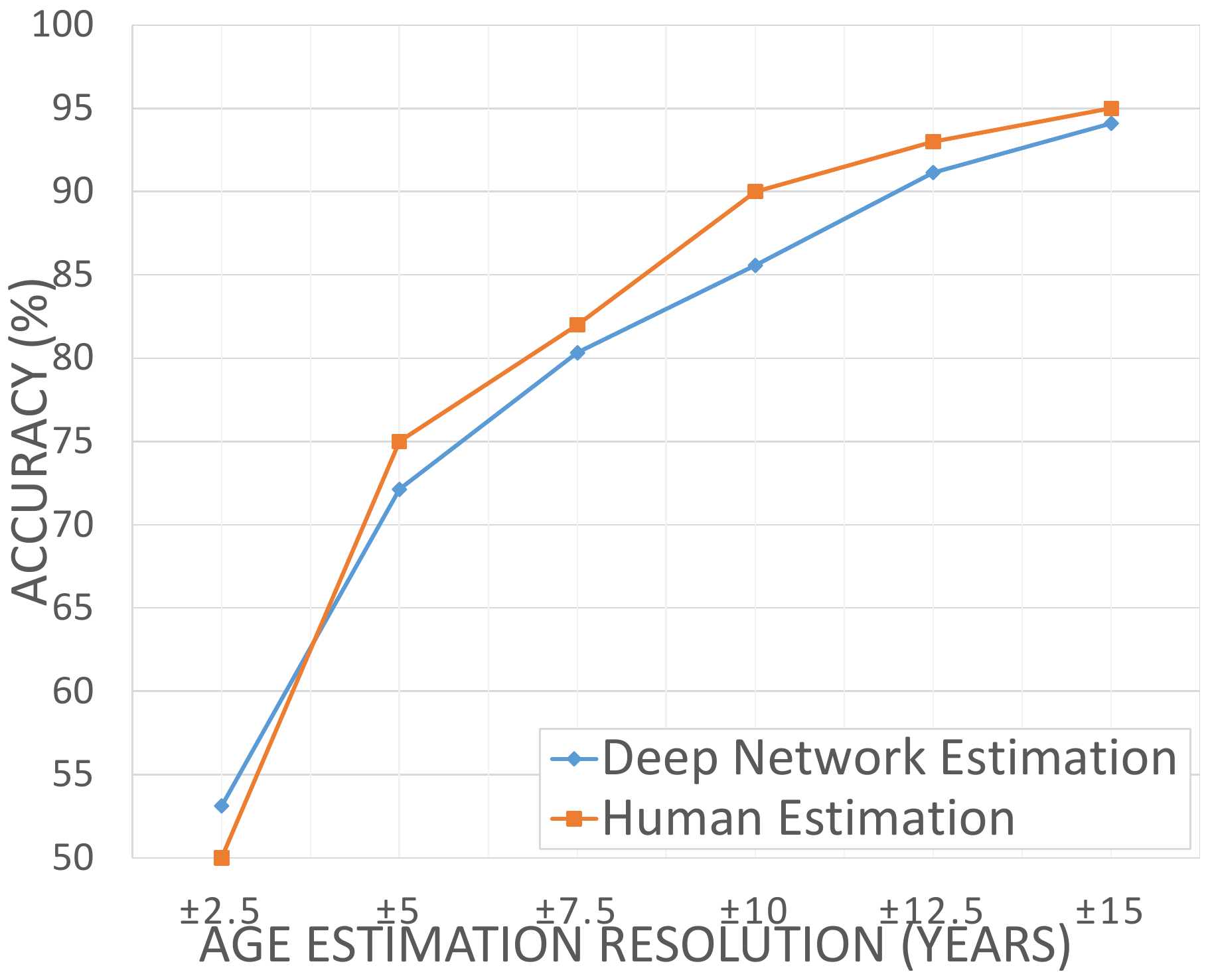}
	   	\caption[Age estimation: humans vs machines]{Age estimation: humans (FG-NET) \cite{han2013age} vs machines (VV Dataset)}
	   	\label{fig:agedlvshumans}
    \end{subfigure} 
    \caption[Age classification on the VV dataset]{Age classification on the VV dataset: The total classification accuracy was 53.12\%/72.13\% for $\pm2.5$ year/$\pm5$ year resolution. The average precision for younger than 50 years age group was 33.3\%/51.7\%.} 
    \label{fig:VVAgeResults}
\end{figure}

Performance of the network on the test set can be seen in Figure \ref{fig:VVAgeResults}: the green squares represent correct classification within $\pm2.5$year resolution, while the orange squares represent correct classification within $\pm5$year resolution. It can be seen that the distribution of age within this dataset is quite skewed towards the age range of [23-27] (refer Figure \ref{fig:VVDatasetHistogram}), and the result of this can be seen in the confusion matrix: there is a bias in the network towards [23-27] age class. Also note that due to the extreme lack of data samples in the above 50 age range, the network performance is severely degraded. This is the main reason for a low average precision while having a high total classification accuracy.

The task of age estimation from faces is something that humans do inherently. It has been observed that age estimation by humans is accurate only up to a range of $\pm4.2$ to $\pm7.4$ \cite{han2013age}. Figure \ref{fig:agedlvshumans} shows the performance of the deep network approach to automated age estimation for the VV dataset vs the age estimation by humans at various resolutions for the FG-NET ageing dataset. 
Due to the similarity in the type of images and the reasonably large size of the FG-NET dataset and the VV dataset, we can assume the performance of humans to be similar on the VV dataset. As can be seen, the performance of the deep network estimation is fairly close that of humans. 

\vspace{-0.2cm}
\subsubsection{Gender Classification}

For this experiment, the deep network was trained on the gender annotated part of the VV dataset by modifying the final softmax output layer to only contain two neurons (for male and female). 


\begin{figure}
	\centering
     \begin{subfigure}{0.435\linewidth}
        \centering
    	\includegraphics[width=\linewidth]{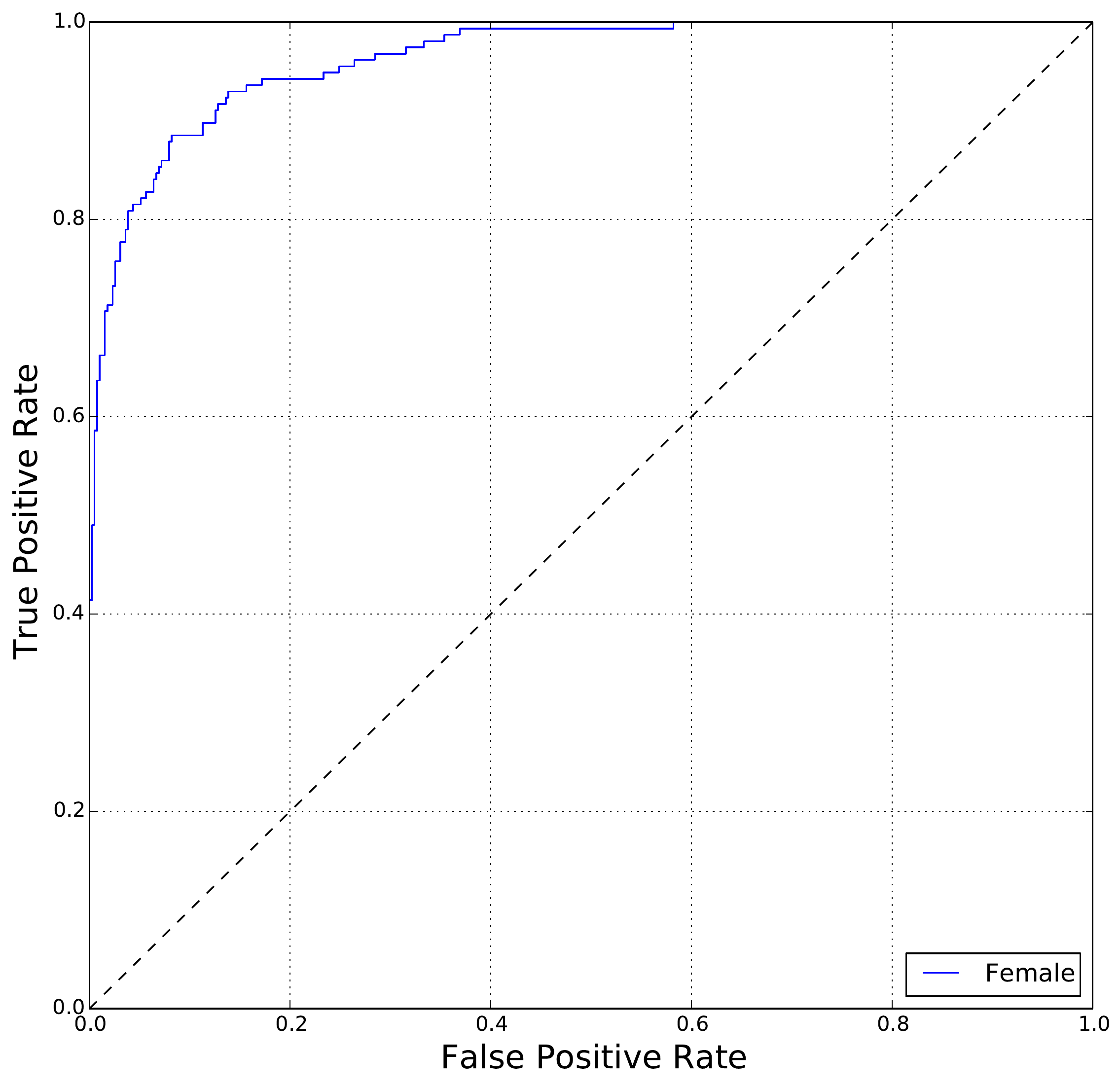}
    	\caption{VV dataset gender classification ROC (female). Class. Acc = 90.68\%, AP = 88.9\%} 
    	\label{fig:VVGenResults}
    	\end{subfigure}
        \begin{subfigure}{0.555\linewidth}
        \centering
           	\includegraphics[width=\linewidth]{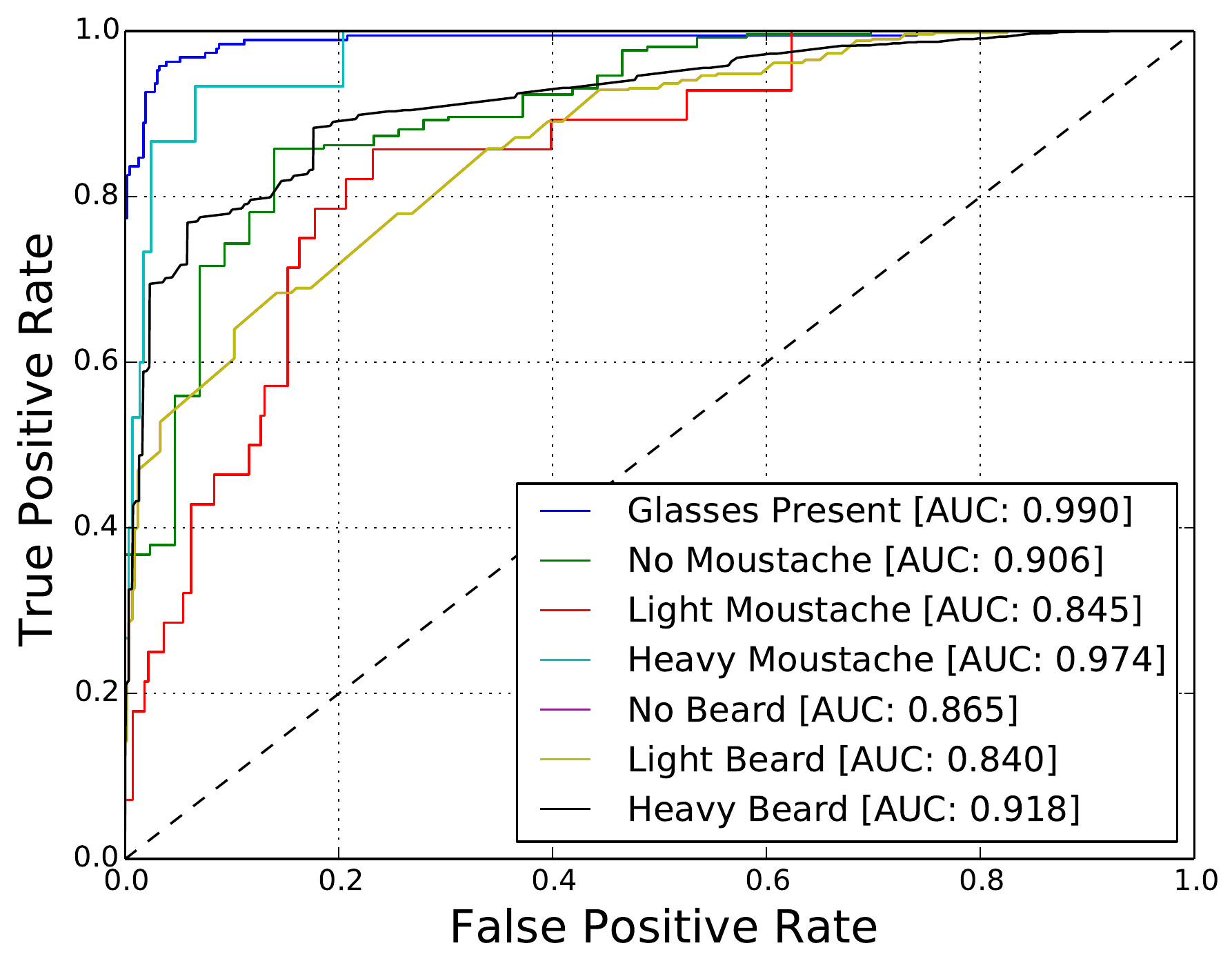}
           	\caption[Network performance for glasses and facial hair detection]{The ROC plot for the network performance for glasses and facial hair detection.}
           	\label{fig:GBMROC}
        \end{subfigure}%
        \caption[GGBM classification on the VV dataset]{Gender and facial hair classification on the VV dataset} 
        \label{fig:VVGGBMResults}
\end{figure}

\begin{figure}
    \centering
   	\includegraphics[width=\linewidth]{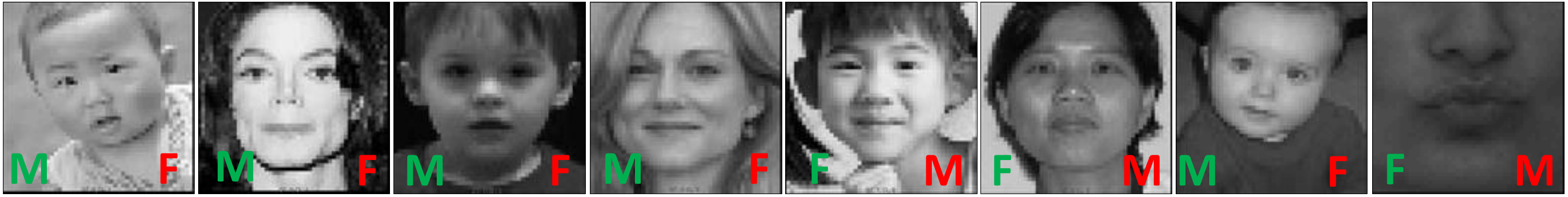}
   	\caption[Gender misclassification examples]{Gender misclassification examples. Legend: Green = Ground Truth, Red = Network classification.} 
   	\label{fig:VVGenErrors}
\end{figure}

The test set performance of the network is plotted in Figure \ref{fig:VVGenResults}.The network is able to correctly classify 90.86\% of the faces in the test set, and the ROC curve shows good discrimination characteristics. Examples of the falsely classified faces can be seen in Figure \ref{fig:VVGenErrors} where a large portion of the misclassified faces are those of young children.
The misclassified images also include difficult to classify faces with non-prominent or mixed gender features (see third from right and second from left in the figure). Lastly, there also exist a small portion of examples with incorrect ground truth (fourth image from left).
Overall, a classification accuracy of above 90\% and the presence of such hard-to-classify faces in the test set suggests that the network performs on a near human level on the task of gender classification.

\vspace{-0.3cm}
\subsubsection{Ethnicity Classification}
The VV dataset contains annotations for five classes of ethnicity, the fifth one being `others' which includes all other non-listed ethnic groups like Middle-Eastern, Latin-American, etc. 


The deep network's final softmax output layer was again altered to fit the annotation by setting the number of neurons to 5.
The network produced a total classification accuracy of 92.24\%, with the average precision for all classes being 61.52\%.
It is also apparent from Figure \ref{fig:VVDatasetHistogram} that the distribution of classes is highly uneven, with the Caucasian and East Asian samples being more abundant than African, South Asian and others.
The effect of this can be seen in the performance of the network: the network performs very well for Caucasian and East Asian faces (above 95\% precision), but the average precision of all classes is only 61.52\%.
Another important point here is that the network does not use the color information in the images, and ethnicity is one of the facial attributes that exhibits a high variance in the color of the skin. 


\subsubsection{Detection of Glasses and Facial Hair}

The network is trained using the same setup as described in the previous experiments, with the final softmax later altered to have only two neurons for the presence of glasses, and 3 neurons for the amount of mustache and amount of beard (none, light, heavy). The performance of the network on the test sets were as follows: 94.52\% accuracy for the presence of glasses, and 88.1\% for the amount of beard and 89.13\% for the amount of mustache. The ROC curves for the classification labels are plotted in Figure \ref{fig:GBMROC}. As can be seen, the network's precision for detecting the presence of glasses and heavy mustache is very high. However, due to the slightly ambiguous definition of light beard class in the dataset, the network does not learn a very precise light beard (and no beard) classifiers.


\vspace{-0.3cm}
\subsubsection{AAM Modelling of Faces using Deep Learning}

The Active Appearance Model (AAM) \cite{van2005model} produces a 3-D model of the face using a compressed representation that encodes the shape and appearance (including texture) of the face together. As briefly mentioned in Section \ref{RelatedWork}, the AAM is conventionally produced by applying PCA directly on the pre-processed pixels of the face image. The shape and appearance parameters within it are encoded as the deviation of a face from the average face. 
Apart from this, it also consists of the pose of the face: the angles made by the normal of the face with respect to the X, Y and Z axes. 
However, as mentioned before in pre=processing sub-section \ref{preprocessing}, the in-plane rotation of the image removes the X-Y plane tilt of the face.
Finally, this annotation is expressed to the network in terms of a compressed vector of AAM parameters plus 2 angles.

\begin{figure}
    \centering
   	\includegraphics[width=0.5\linewidth]{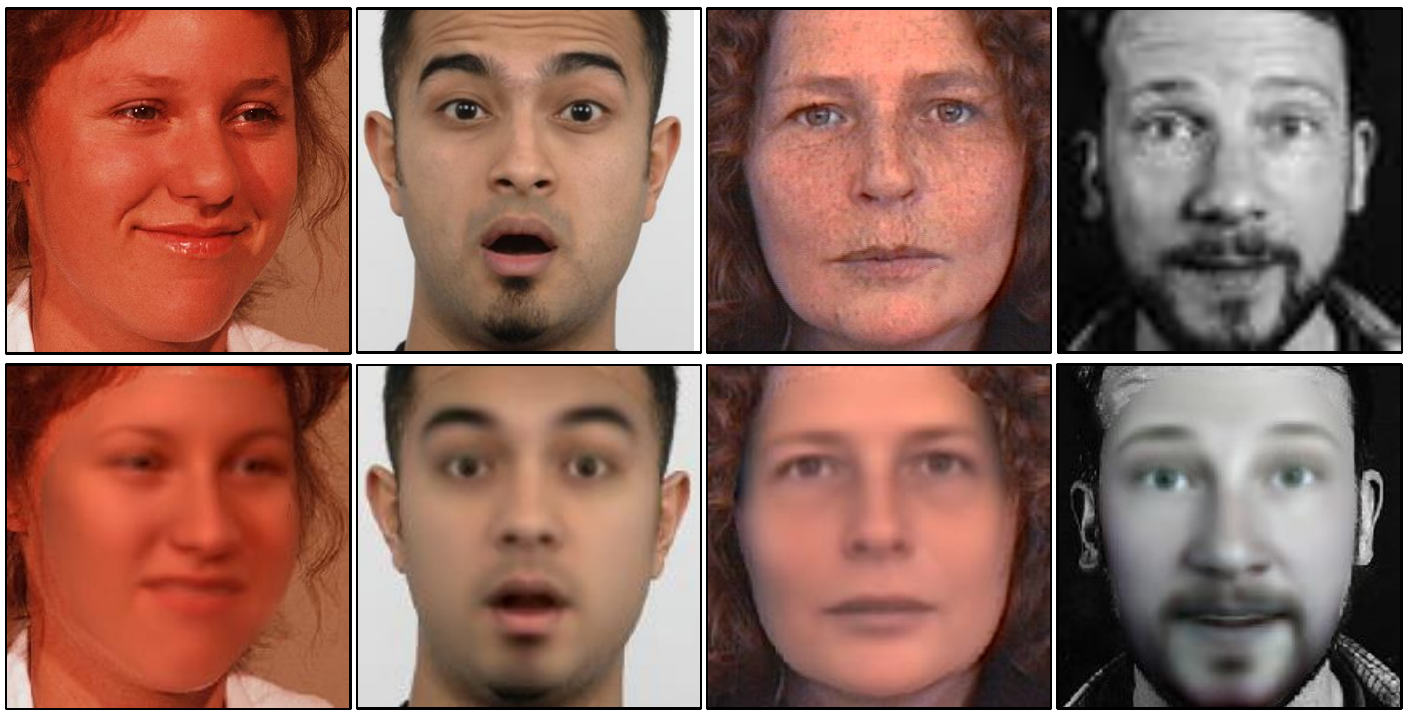}
   	\caption[AAM based synthetic faces]{Examples of synthetic faces used for training the network (bottom), and their source images (top). The synthetic faces are obtained using AAM.} 
   	\label{fig:synthetic}
\end{figure} 

The network was trained on synthetic faces and their corresponding Active Appearance Models generated by the conventional face modelling method described above (see Figure \ref{fig:synthetic}). The reason for using synthetic faces instead of real faces is that because these synthetic faces are generated by their corresponding AAM, the modelling error between the face and the AAM vector is zero. 

\begin{figure}
	\centering
	\includegraphics[width=0.85\linewidth]{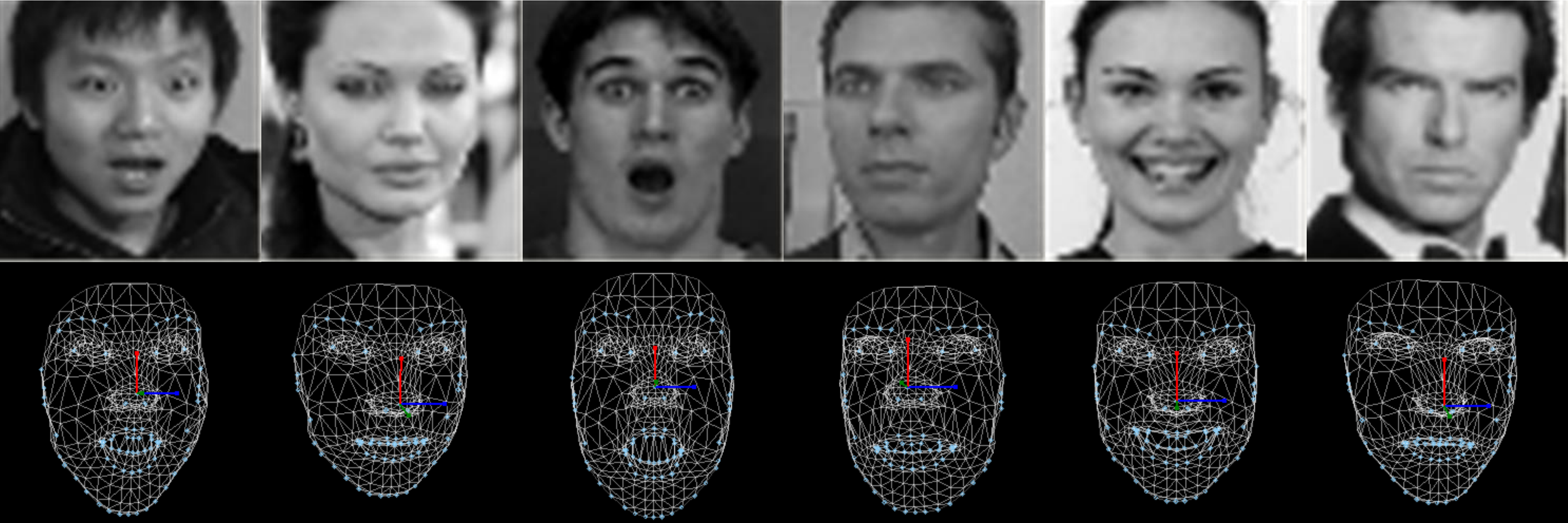}
	\caption{Face models generated by the deep network.}
    \label{fig:aams}
\end{figure}


A cosine similarity score of 0.768 is obtained with the ground truth when tested on real faces, and a similarity score of 0.862 when tested on synthetic faces. Moreover, the pose estimation for the test faces produced an average error of $2.92^\circ$/$1.89^\circ$ in the Y/X axes for real faces, and $2.23^\circ$/$1.66^\circ$ in the Y/X axes for synthetic faces. 
As can be seen in Figure \ref{fig:aams}, the generated face models resemble the real faces in terms of shape and pose quite well.


\vspace{-0.2cm}
\subsubsection{Relation between High-level Concepts and Low-level Descriptors}
In all the experiments explained above, the deep network was required to be trained on specific annotation-image pairs for the given task. However, it can be argued that high-level features in faces (like age, emotions) are just combinations of certain low-level features in faces (like eye-edges, lip-curl), and many of these low-level features can be common among different high-level feature descriptions. 

\vspace{-3ex}
\paragraph{Similarity in First-Layer Weights}
\noindent To study the above mentioned argument, this sub-section compares the low-level descriptors that combine to form different higher level concepts. Careful observation of the first convolution layer reveals a similarity in the general pattern of the weights of the feature-maps. 

\begin{figure}
	\centering
	\includegraphics[width=0.95\linewidth]{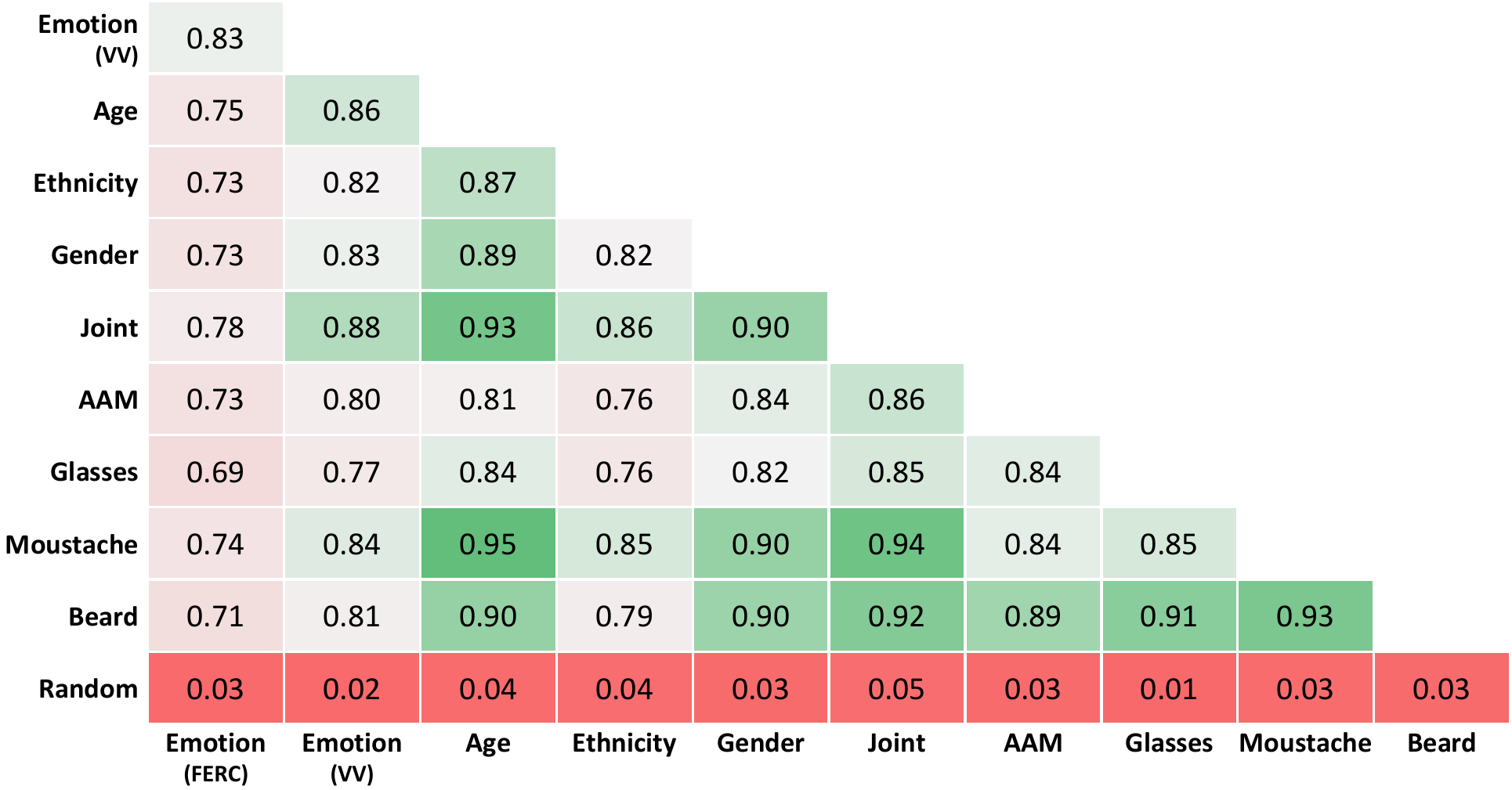}
	\caption[Similarity in learned weights]{Cosine inter-similarity scores for first-layer weights learnt by the network for different tasks.}
 \label{fig:weightSim}
\end{figure}

Figure \ref{fig:weightSim} illustrates a heat-table of cosine similarity scores between the first layer weights of networks trained for different high level feature recognition (classification task). Certain patterns can be observed in these results. Weights for facial-hair and gender are very similar to those of age: this could be related to the fact that all females and all young children have no facial hair, and hence the presence or absence of facial-hair can be a good indicator of the person's gender and age. This could also be the reason for a high similarity between age and gender weights. Weights of emotions are dissimilar from the weights of age, ethnicity, gender, glasses and facial-hair, as these factors do not influence the facial expressions of a person. On the other hand, weights of joint classification (explained later) are similar to all the other tasks since joint classification involves the classification of all those features.


In general, it could be observed that the given task seems to have a strong effect on the lower-level descriptors. A higher correlation in the weights for similar visual tasks is also observed, while it is seen that visually dissimilar tasks exhibit a lower correlation. 


\vspace{-3ex}
\paragraph{Joint Classification Experiment}
In order to exploit the low-level similarity observations from the previous sub-section, a single network needs to be trained to jointly classify multiple non-exclusive facial features.
In order to achieve this, the different annotations per image in the VV Dataset are combined into one single set of image -- annotation pair, where the annotations are represented by 37 non-exclusive class labels. 

The difference between the performance of the joint classification network and individual networks was found to be very small: on average 1.84\% ([0.91\% - 4.71\%]) lower than the accuracy of individual networks. 
This suggests that the deep network is capable of learning to classify multiple semantic features in faces in a joint manner. 
\vspace{-0.25cm}
\section{Conclusion}
\label{Conclusion}
\vspace{-0.1cm}
In this paper, a deep learning based approach has been demonstrated for the task of semantic facial feature recognition. This approach is primarily based on the use of convolutional neural networks on two dimensional pre-processed and aligned images of faces. A study exploring the effects of network hyper-parameters on the classification performance has been conducted, leading to estimation of the near-optimal configuration of the network. The study suggests that a deep convolutional network based approach is naturally well suited for the task of image based facial expression recognition. It is shown that addition of deterministic pre-processing and alignment steps for the input data greatly aids in improving the performance. Such a deep network can easily be adapted to the tasks of recognizing additional semantic features. Experimental results have shown near-human performance. However, the discrimination power of deep networks are highly dependent on the distribution and quality of the training data. The relation between the high-level semantic features and low-level descriptors has also been studied. Specific intuitive similarities have been observed between the low-level descriptors for different tasks. Use of this commonality among low-level descriptors is demonstrated by training a single network to jointly classify multiple semantic facial features. 
Finally, a novel scheme for training deep networks to generate complete 3-D Active Appearance Models of faces from 2-D images has been shown. To our best knowledge this is the first time deep networks is used to predict a compressed image representation and this task has also been successfully achieved by the network. 



{\small
\bibliographystyle{unsrt}
\bibliography{Bibliography}
}

\end{document}